%% file: main.tex
\documentclass[runningheads]{llncs}

\usepackage{iciap}

\usepackage{iciapabbrv}

\usepackage{graphicx}
\usepackage{booktabs}

\usepackage[accsupp]{axessibility}  

\input{preamble}

\usepackage{hyperref}

\usepackage{orcidlink}

\begin{document}

\title{Intrinsic Training Signals for \\ Federated Learning Aggregation}

\titlerunning{\methname}

\author{Cosimo Fiorini\orcidlink{0009-0007-4083-0390} \and
Matteo Mosconi\orcidlink{0009-0008-1989-5779} \and
Pietro Buzzega\orcidlink{0000-0002-6516-6373} \and
\\Riccardo Salami\orcidlink{0009-0002-0704-5810} \and
Simone Calderara\orcidlink{0000-0001-9056-1538}}

\authorrunning{C.~Fiorini \textit{et al.}}

\institute{AImageLab - University of Modena and Reggio Emilia, Italy \email{name.surname@unimore.it}}

\maketitle

\input{sections/0_abstract}
\input{sections/1_introduction}
\input{sections/2_related}
\input{sections/3_background}
\input{sections/4_method}
\input{sections/5_experiments}
\input{sections/6_conclusions}

%
%
\bibliographystyle{splncs04}
\bibliography{bibliography}

\newpage
\input{supplementary/0_supplementary}
\input{supplementary/1_shapvalues}
\input{supplementary/2_varimportance}
\input{supplementary/3_a_posteriori}

\end{document}

%% file: preamble.tex
\usepackage{xspace}
\usepackage{amssymb}
\usepackage{amsmath}
\usepackage{dsfont}
\usepackage{multirow}
\usepackage{float}
\usepackage{lipsum}
\usepackage{nicefrac}
\usepackage{circledsteps}
\usepackage{enumitem}
\usepackage{colortbl}
\usepackage{xcolor}
\usepackage{siunitx}
\usepackage{breqn}
\usepackage{lipsum}
\usepackage{caption}
\usepackage{duckuments}
\usepackage{tikzducks}
\usepackage{tablefootnote}
\usepackage{pifont} 
\usepackage{makecell}
\usepackage{tikz}
\usepackage{algorithm}
\usepackage{algpseudocode} 
\usepackage{tcolorbox}
\usepackage{soul}

\definecolor{lightgray}{gray}{0.95}
\definecolor{midgray}{gray}{0.55}
\definecolor{steelblue}{HTML}{4D82B7}
\definecolor{davysgrey}{rgb}{0.33, 0.33, 0.33}
\definecolor{LightCyan}{rgb}{0.88,1,1}

\usepackage[first=0, last=9]{lcg}

\newcommand{\cmark}{\ding{51}}%
\newcommand{\xmark}{\ding{55}}%

\newcommand{\Star}[1]{#1\ensuremath{^*}\kern-\scriptspace}

\newcommand{\tit}[1]{\smallbreak\noindent\textbf{#1}}

\newcommand{\plus}{\texttt{+}}

\newcommand{\methname}{LIVAR\xspace}

\renewcommand{\algorithmiccomment}[1]{\bgroup\hfill $\triangleright$ ~#1\egroup}


%% file: sections/0_abstract.tex
\begin{abstract}
Federated Learning (FL) enables collaborative model training across distributed clients while preserving data privacy. While existing approaches for aggregating client-specific classification heads and adapted backbone parameters require architectural modifications or loss function changes, our method uniquely leverages intrinsic training signals already available during standard optimization. We present \methname  (Layer Importance and VARiance-based merging), which introduces: \textit{i)} a variance-weighted classifier aggregation scheme using naturally emergent feature statistics, and \textit{ii)} an explainability-driven LoRA merging technique based on SHAP analysis of existing update parameter patterns. Without any architectural overhead, \methname  achieves state-of-the-art performance on multiple benchmarks while maintaining seamless integration with existing FL methods. This work demonstrates that effective model merging can be achieved solely through existing training signals, establishing a new paradigm for efficient federated model aggregation. The code is available at \url{https://github.com/aimagelab/fed-mammoth}.
\keywords{Federated Learning \and Model Merging \and Explainability}
\end{abstract}

%% file: sections/1_introduction.tex
\section{Introduction}
The limited ability of deep models to effectively learn from sparse and distributed real-world data has necessitated the development of Federated Learning (FL), a paradigm in which clients train on their local data distributions and share model updates with a central server. The server aggregates these updates without accessing raw local data, thereby preserving privacy. FL has found practical applications in diverse domains, such as manufacturing, where factories collaborate to build a unified model without sharing proprietary sensor data~\cite{pruckovskaja2023federated}, and healthcare, where hospitals cannot exchange sensitive patient information~\cite{sheller2020federated}.
In recent years, Parameter-Efficient Fine-Tuning (PEFT)~\cite{fu2023effectiveness} has emerged as a training paradigm well-suited to FL. PEFT techniques leverage pre-trained architectures~\cite{ridnik2021imagenet} to reduce the number of trainable parameters while maintaining performance comparable to full fine-tuning. Beyond computational efficiency, PEFT is particularly advantageous in FL, as it minimizes communication overhead by transmitting fewer parameters.

Recent studies employing pre-trained architectures highlight distinct contributions of fine-tuning the backbone versus the final classification layer, with evidence suggesting adapting the classifier plays a more critical role in model performance~\cite{zhang2023slca}. Consequently, some FL methods employ specialized strategies for tuning the classifier~\cite{luo2021no} while aggregating backbone parameters via simple averaging~\cite{mcmahan2017communication}. Others utilize PEFT techniques, like LoRA~\cite{hu2021lora}, to fine-tune the backbone~\cite{sun2024improving} and subsequently merge updated vectors. Research explored various approaches for determining module aggregation coefficients~\cite{yadav2024ties,yang2024adamerging}, including learnable weighting mechanisms~\cite{wu2024mixture} and closed-form solutions~\cite{salami2025closed}.

Alternatively, other approaches employ separate training and aggregation solutions for the backbone and classification head~\cite{salami2024federated}. While we consider this decoupled strategy superior, we also advocate for exploring more lightweight solutions for both classifier head merging and backbone module aggregation. We argue that leveraging naturally available training signals can establish both efficient and effective Federated Learning baselines.

Building upon these foundations, we present \methname, a novel Federated Learning framework that implements a decoupled approach for aggregating both backbone parameters and classification heads. For classifier merging, we propose variance-weighted heads, where each client’s class-specific head is weighted by the variance of features from correctly predicted training samples of that class. Intuitively, higher variance indicates richer class-specific information, and this measure is obtained effortlessly during local training.

For backbone adaptation, \methname employs LoRA modules, integrating them with a novel explainability-driven aggregation scheme. We're the first to propose applying SHAP values~\cite{lundberg2017unified} to determine optimal merging coefficients for federated LoRA modules. Aggregation weights are derived from layer-wise gradients and parameter update patterns during client training; larger cumulative updates signify more substantial adaptations of specific network components. These naturally emergent signals -- available without architectural modifications -- serve as proxies for each client's contribution to particular model parameters. To implement this, we leverage a proxy dataset to construct an explainability curve that directly maps these adaptation patterns to their optimal merging coefficients.

We extensively evaluate \methname on in-distribution and out-of-distribution datasets \wrt to pre-training. Our method achieves state-of-the-art performance in all in-distribution settings and remains competitive otherwise. Furthermore, we demonstrate that our classifier weighting scheme remains compatible with existing head-specific strategies, enabling seamless integration with these approaches. Beyond FL, our work highlights the broader potential of explainability-driven module merging, opening new research directions in model aggregation.

In summary, our contributions are as follows:
\begin{itemize}[label=$\bullet$]
\item We propose a variance-weighted aggregation strategy for classification heads.
\item We conceive explainability-driven merging coefficients for the federated aggregation of LoRA modules.
\item We propose \methname, achieving state-of-the-art results across multiple benchmarks while maintaining seamless integration with existing approaches.
\end{itemize}

%% file: sections/2_related.tex
\section{Related Works}
\textbf{Federated Learning} addresses the challenges of training models in a distributed environment, where local data are private and cannot be exchanged. Federated Averaging (FedAvg)~\cite{mcmahan2017communication} established the foundation by averaging client model weights, yet its simplicity makes it vulnerable to highly heterogeneous data streams, which inherently introduce local client divergence. To overcome this challenge, techniques such as Scaffold~\cite{karimireddy2020scaffold} introduce control variates that compensate for local updates, thereby reducing client drift. Other approaches include FedProx~\cite{li2020federated}, which incorporates a proximal term to constrain the deviation between local and global models, and MOON~\cite{li2021model}, which employs contrastive loss to align the representations learned by local and global models. Similarly, CCVR~\cite{luo2021no} enhances the classifier on the server side by rebalancing it with synthetic features, thereby stabilizing updates across heterogeneous clients and promoting convergence. FedProto~\cite{tan2021fedproto} leverages prototypical representations to regularize the local training process, enabling the maintenance of class-specific information, while GradMA~\cite{luo2023gradma} uses a gradient-memory mechanism to guide local update processes by incorporating historical gradient information, thereby stabilizing update and accelerating model convergence. 

\textbf{Model Merging} methodologies have emerged as powerful alternatives to conventional weight averaging in FL systems. Fisher Averaging~\cite{matena2022merging} leverages the Fisher Information Matrix to weight client contributions, thus prioritizing more informative updates during aggregation. RegMean~\cite{jin2022dataless} presents a closed-form solution for merging model weights, enabling the identification of a global model that best approximates the responses of the clients, demonstrating strong potential for federated settings. Following the introduction of LoRA (Low-Rank Adaptation)~\cite{hu2021lora}, a growing body of research has emerged to address the challenge of optimal coefficient selection for module merging~\cite{huang2023lorahub,yadav2024ties,yang2024adamerging,wu2024mixture}, including several approaches specifically designed for Federated Learning scenarios~\cite{salami2025closed}. While these methods achieve promising results, they commonly rely on auxiliary architectural elements, computationally expensive procedures, or customized loss functions. In contrast, our methodology introduces a more streamlined solution that utilizes naturally emerging training signals to effectively aggregate both the LoRA-adapted backbone and the classification head without requiring additional components.

%% file: sections/3_background.tex
\section{Preliminaries}
\tit{LoRA.} Low-Rank Adaptation~\cite{hu2021lora} is an efficient parameter adaptation technique that significantly reduces the number of trainable parameters in deep neural networks. For a given layer with weight matrix $W \in \mathbb{R}^{d \times k}$, LoRA decomposes the weight update $\Delta W$ into two trainable low-rank matrices: $B \in \mathbb{R}^{d \times r}$ and $A \in \mathbb{R}^{r \times k}$, where the rank $r\ll min(d,k)$. The adapted weights are computed as:
\begin{equation}
W' = W + BA
\end{equation}
During adaptation, the original weights $W$ remain frozen while only the low-rank matrices $A$ and $B$ are updated. This decomposition reduces the number of trainable parameters from $d \times k$ to $r \times (d + k)$, resulting in significant memory savings compared to full fine-tuning.

\tit{Problem Setting.} We address a $C$-class classification problem in Federated Learning where data are distributed non-IID across $M$ clients. Let $\mathcal{D}_{m}$ represent the data partition for client $m$. The training process consists of sequential communication rounds; at each round, clients perform local optimization for a specified number of epochs, and then the server aggregates the optimized parameters before redistributing them to clients. Let $f_{\theta}$ denote the global model parameterized by $\theta$.
The training procedure aims to find the optimal parameter set $\theta$ that minimizes the loss across the entire dataset:
\begin{equation}
\underset{\theta}{\text{minimize}} \quad \frac{1}{M} \sum_{m=1}^M \mathbb{E}_{(x,y)\sim \mathcal{D}_m}\mathcal{L}(f_{\theta}(x), y)
\end{equation}

%% file: sections/4_method.tex
\section{\methname}
\input{figures/main_fig}
We present \methname, a Federated Learning framework with distinct aggregation strategies for the backbone and the classification head. Each client $m$ maintains parameters $\theta_m = \{W_m, H_m\}$, where $W_m$ represents the backbone weights and $H_m$ denotes the classification head.

For the backbone, we employ Low-Rank Adaptation with our novel \emph{Proxy-Calibrated Layer Importance Weighting} mechanism, using explainability-driven coefficients to merge client updates. The classification head uses \emph{data-variance-aware} merging that automatically adapts to client data heterogeneity, ensuring robustness under non-IID distributions.
The dual merging strategy is depicted in \Cref{fig:model}.

\subsection{Proxy-Calibrated Layer Importance Weighting}
\label{sec:proxy_calibration}
\input{figures/shap_mainmethod}

Each client $m$ learns $L$ low-rank adaptations $\Delta W_m = \{\Delta W_m^1, \ldots, \Delta W_m^L\}$, where each adaptation $\Delta W_m^l = B_m^lA_m^l$ follows the standard LoRA decomposition. The server aggregates these client updates through a layer-wise weighted combination:
\begin{equation}
\Delta W^l = \sum_{m=1}^M \alpha_m^l \cdot \Delta W_m^l
\label{eq:lora_weighting}
\end{equation}
where $\alpha_m^l$ represents the layer-specific merging coefficients that constitute the target of our estimation procedure.

These $\alpha_m^l$ coefficients are derived through a three-stage process. First, we establish a ground truth reference using a proxy dataset (\textit{i.e.}, Tiny-ImageNet~\cite{le2015tiny}) by training five federated clients in parallel with a centralized model trained on the full dataset. This process yields the reference parameters $\Delta W_{\text{joint}}^l$ of the centralized model. The fundamental premise is that optimal aggregation should minimize the distance between federated combinations and centralized training outcomes. Formally, for each layer $l$, we solve the constrained regression problem:
\begin{equation}
\underset{\lambda_m^l}{\text{minimize}} \quad \|\Delta W_{joint}^l - \sum_{m=1}^5 \lambda_m^l \Delta W_m^l \|_2^2 \quad \text{subject to} \quad \lambda_m^l \geq 0
\end{equation}
Second, we generalize this solution by establishing a mapping from trainable parameters to optimal coefficients. Our method builds on the key insight that a layer’s contribution to loss variation effectively captures its importance during server-side aggregation, providing a reliable basis for estimating the optimal merging coefficients $\lambda_{m}^l$. Following~\cite{zenke2017continual}, we compute layer importance measures $\omega_m^{l_A}$ and $\omega_m^{l_B}$ as the running sums of Hadamard products between gradients and parameter updates for matrices $A_m^l$ and $B_m^l$ respectively. To find the relationship between layer importance scores $\{\omega_m^{l_A}, \omega_m^{l_B}\}$ and actual coefficients $\lambda_m^l$, we train a Random Forest regressor~\cite{breiman2001random} to predict $\lambda_m^l$ from the importance signals. This enables a robust, explainable, and efficient estimation of coefficients on new datasets, without relying on a centralized (joint) reference model.

Third, we employ SHAP analysis~\cite{lundberg2017unified} to interpret the trained model (see \Cref{fig:shapmain}), revealing distinct roles of the $A$ and $B$ matrices – the former initialized via Kaiming uniform distribution~\cite{he2015delving} to explore directions, the latter initialized to zeros to control direction utilization. The SHAP values are discretized through a quantile-based function $G_{SHAP}$ to yield the final coefficients:
\begin{equation}
\alpha_m^l = G_{SHAP}(\omega_m^{l_A}, \omega_m^{l_B})
\label{eq:alpha_def}
\end{equation}
Additional implementation details regarding the quantile-based discretization procedure are provided in \Cref{sec:calbrationdetails}.

\input{algorithms/main_algo}

\tit{Beyond the proxy dataset.} The function $G_{SHAP}(\cdot)$, which estimates merging coefficients from importance signals $\omega_m^{l_A}$ and $\omega_m^{l_B}$ observed during training, enables federated aggregation across arbitrary datasets. Our quantile-based discretization ensures robust generalization beyond the proxy dataset used for calibration. We validate this generalization capability through a posteriori analysis of SHAP value distributions across multiple benchmarks in \Cref{sec:a_posteriori}.

For server-side aggregation, the coefficients $\alpha_m^l$ computed via \Cref{eq:alpha_def} are applied to combine client updates $\Delta W_m^l$, as in \Cref{eq:lora_weighting}, with each client's importance measures $\omega_m^{l_A}$ and $\omega_m^{l_B}$ derived from their local training process.

\subsection{Head Class-specific Variance Weighting}
The classification head aggregation component of \methname operates on the principle that feature-space variance correlates with class representation quality. At the end of client-side training, we compute the mean variance $\sigma_{m}^{(c)}$ of pre-logits features for correctly classified examples of each class $c$, serving as a proxy for how comprehensively each client represents the class.

At the server, aggregation weights are derived from normalized variance contributions across clients. For each class $c$, the server first computes the total variance $\Sigma^{(c)} = \sum_m \sigma_m^{(c)}$ across all clients, then performs the weighted aggregation as:
\begin{equation}
H^{(c)} = \sum_m \frac{\sigma_m^{(c)}}{\Sigma^{(c)}} H_m^{(c)}
\label{eq:head_weighting}
\end{equation}
This variance-weighted scheme prioritizes classification heads whose training data exhibits broader coverage of each class's feature distribution. 
The complete training procedure is formally specified in \Cref{algo:model}.

%% file: figures/main_fig.tex
\begin{figure}[t]
    \centering
    \includegraphics[width=1\textwidth]{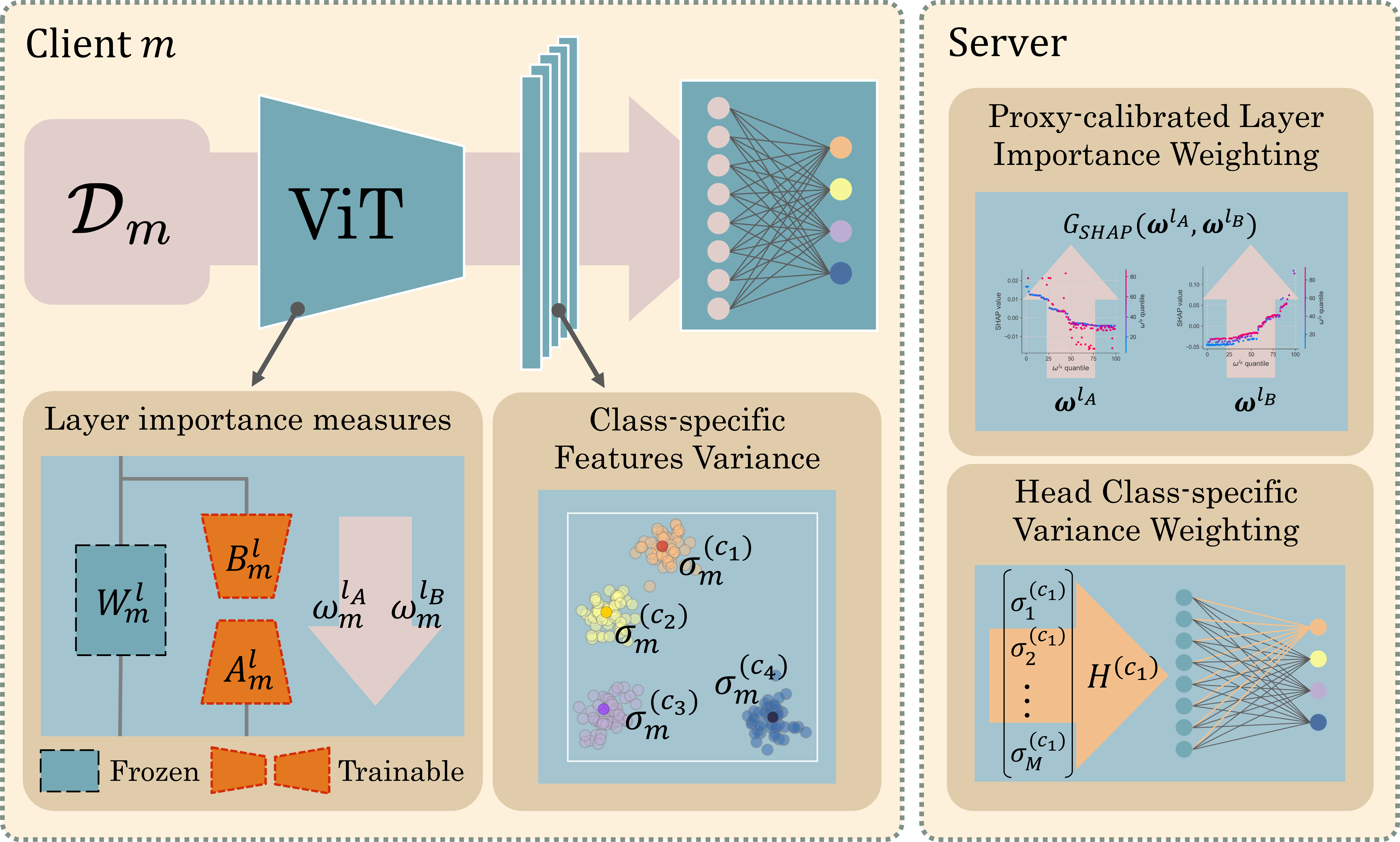}
    \caption{\methname employs a dual weighting strategy architecture. The left diagram depicts client-side information collection during training, the right diagram shows the distinct weighting strategies used on the server side.}
    \label{fig:model}
\end{figure}

%% file: figures/shap_mainmethod.tex
\begin{figure}[t]
    \centering
    \includegraphics[width=\columnwidth]{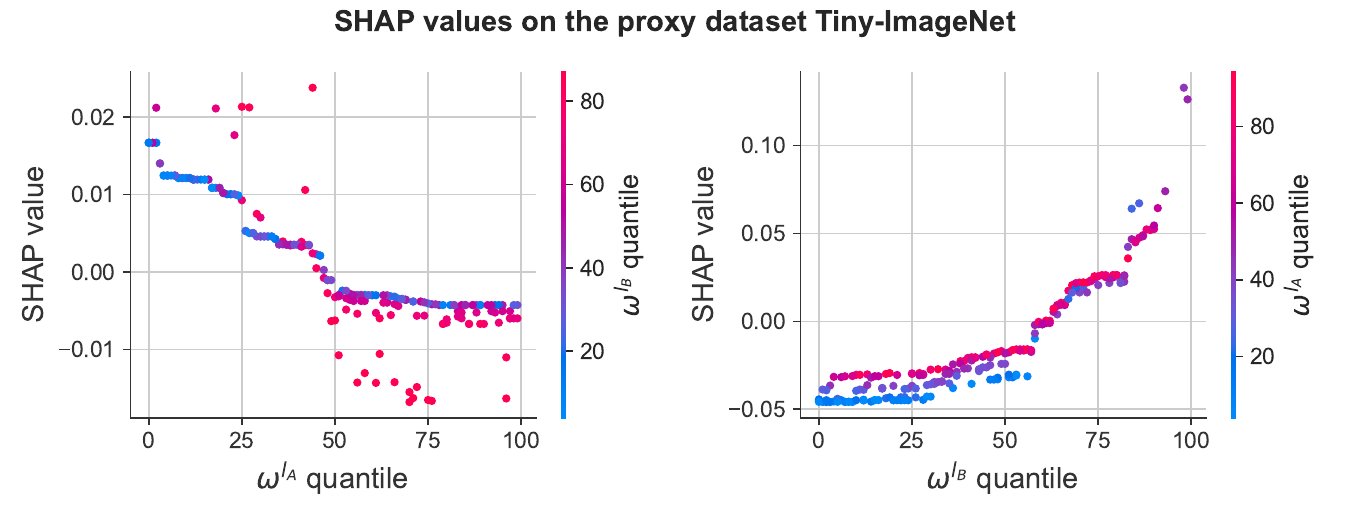}
    \caption{SHAP values represent the impact of $\omega_{m}^{l_A}$ and $\omega_{m}^{l_B}$ features in the prediction of optimal merging coefficients $\lambda_m^l$. Values on x axis are expressed in quantiles with respect to the values of all clients and all layers.}
    \label{fig:shapmain}
\end{figure}

%% file: algorithms/main_algo.tex
\begin{algorithm}[t]
\caption{\methname training for a single round}
\label{algo:model}
\begin{algorithmic}

\State \textbf{Input:} $M$ clients, $L$ LoRA layers, $G_{SHAP}(\cdot)$ proxy-calibrated function, $C$ classes.

\State \underline{Client side:}
\For{client $m \in \{1, \ldots, M\}$}

    \State $\Delta W_m \gets (training)$  \algorithmiccomment{Collect $(\omega_m^{l_A},\omega_m^{l_B})$ during training}
    
    \State Collect $\sigma_{m}^{(c)}$ over well classified examples

    \State Send $(\omega_m^{l_A},\omega_m^{l_B})$ and $\sigma_{m}^{(c)}$ to the server
    
\EndFor

\State \underline{Server side:}
\For{layer $l \in \{1, \ldots, L\}$}

    \State $\alpha_m^l \gets$ use \Cref{eq:alpha_def} 
    
    \State $\Delta W^l \gets$ use \Cref{eq:lora_weighting} \algorithmiccomment{Proxy-Calibrated Layer Importance Weighting}
    
\EndFor

\For{class $c \in  \{1, \dots, C\}$}

    \State $\Sigma^{(c)} = \sum_m(\sigma_{m}^{(c)})$
    
    \State $H^{(c)} \gets$ use \Cref{eq:head_weighting} \algorithmiccomment{Head Class-Specific Variance Weighting}
    
\EndFor

\end{algorithmic}
\end{algorithm}

%% file: sections/5_experiments.tex
\section{Experimental Studies}
\input{tables/in_distribution}

\subsection{Evaluation Settings}
We conducted a comprehensive evaluation of \methname on both in-distribution datasets (CIFAR-100~\cite{krizhevsky2009learning}, ImageNet-R~\cite{hendrycks2021many}) and out-of-distribution benchmarks (EuroSAT~\cite{helber2019eurosat}, CUB-200~\cite{wah2011caltech}) relative to the pre-training domain. Data was distributed across $10$ clients using a \textit{distribution-based} label imbalance setting governed by a Dirichlet distribution with parameter $\beta$. We evaluated all methods under three distribution-based imbalance scenarios: $\beta \in \{0.2, 0.5, 1.0\}$, where lower values indicate greater data heterogeneity. For all experiments, we evaluated model performance using a global test set.

We benchmark \methname against five state-of-the-art approaches spanning complementary research directions: FedAvg~\cite{mcmahan2017communication}, CCVR~\cite{luo2021no}, and FedProto~\cite{tan2021fedproto} represent federated learning methods, while RegMean~\cite{jin2022dataless} and Fisher Merging~\cite{matena2022merging} constitute established model merging techniques. Our implementation leverages a ViT-B/16~\cite{dosovitskiy2020image} backbone pre-trained on ImageNet-21K~\cite{ridnik2021imagenet} across all methods. For \methname, we applied LoRA adaptation to all \textit{qkv}, \textit{mlp}, and \textit{multi-head attention} projection layers. For all datasets, we standardized the training protocol with $5$ communication rounds, each comprising $5$ local epochs.

\subsection{Results of \methname}
We report comparative results in terms of average accuracy across all evaluated methods.

\tit{In distribution.} As reported in \Cref{tab:main_in}, for both in-distribution datasets, we observe consistent trends: \methname achieves state-of-the-art results across all $\beta$ settings, while CCVR consistently ranks second. For both datasets, higher degrees of distribution imbalance yield more pronounced differences in performance -- while \methname and CCVR maintain their leading performance, the model merging approaches follow closely, and other methods fail to remain robust under these harder conditions.

\input{tables/out_distribution}

\tit{Out of distribution.} \Cref{tab:main_out} shows that the EuroSAT dataset proves less informative for evaluating Federated Learning with ViT-based methods, as all models achieve similar accuracy levels -- likely due to the dataset’s limited $10$-class structure. In contrast, CUB-200 presents a more challenging benchmark where \methname and CCVR significantly outperform competing approaches. \methname achieves state-of-the-art results for moderate data imbalance ($\beta \in \{1, 0.5\}$), with CCVR consistently ranking second. However, under extreme imbalance ($\beta=0.2$), CCVR surpasses \methname, demonstrating the advantages of its classifier rebalancing strategy over our variance-based approach.

\tit{Method composability.} We investigate \methname's compatibility with existing federated learning techniques through integration with CCVR on CUB-200 -- a dataset representing a significant domain shift from ViT's pre-training distribution. While CCVR alone outperforms \methname in the extreme imbalance setting ($\beta=0.2$), we demonstrate their complementary strengths when combined: applying \methname's layer-specific aggregation followed by CCVR's server-side calibration with synthetic data yields superior performance to either method in isolation. As shown in \Cref{tab:composability}, this synergy highlights how \methname's feature variance head weighting provides foundational benefits that enhance subsequent refinement techniques like CCVR. This composability represents a significant advantage, as it allows practitioners to combine multiple techniques to address the unique challenges of different federated learning scenarios.
\input{tables/ablations}

\subsection{Ablation Study on \methname Components}
\label{sec:ablations}
We systematically evaluate \methname's components by measuring Average Accuracy and standard deviation on CIFAR-100 with extreme label imbalance ($\beta=0.05$). Starting with a baseline of FedAvg-aggregated ViT-B/16 parameters -- including both LoRA residuals $\Delta W^l$ and the classification head $H$ -- \Cref{tab:ablation_components} demonstrates progressive improvements through incremental enhancements. The Proxy-Calibrated Layer Importance Weighting ($\alpha^l$) achieves statistically significant accuracy gains beyond the standard deviation across five runs (reported in parentheses in \Cref{tab:ablation_components}). Subsequent incorporation of Head Class-Specific Variance Weighting ($\sigma^{(c)}$) delivers substantially greater improvement, corroborating the classification head's critical role. The full \methname framework combines these components synergistically to achieve peak performance.

\subsection{A Posteriori Analysis of SHAP Results}
\label{sec:a_posteriori}
\input{figures/shap_imagenetr}

To verify the generalizability of our proxy-calibrated SHAP analysis explained in \Cref{sec:proxy_calibration}, we repeated the calibration procedure across all four datasets: CIFAR-100, ImageNet-R, EuroSAT, and CUB-200. Importantly, this analysis does not involve recalibrating the $G_{SHAP}(\cdot)$ function for each specific dataset -- since such per-dataset calibration would be impossible in a federated setting where the joint data distribution is inaccessible -- but instead demonstrates that the original proxy-based calibration effectively generalizes to different data domains.

Following the same methodology that produced the results in \Cref{fig:shapmain}, we generated comparable analyses for each dataset. As shown in \Cref{fig:shapimgr} for ImageNet-R, the SHAP value trends remain consistent: $\omega^{l_A}$ exhibits decreasing or stable influence with increasing feature values, while $\omega^{l_B}$ maintains its clear increasing trend across all datasets. These consistent patterns (fully documented in \Cref{sec:omega}) confirm the proxy dataset's representativeness, despite variations in specific impact magnitudes.

The observed consistency, coupled with the variations in absolute values, provides empirical justification for our $\omega^l$ discretization approach. This design choice ensures \methname's robustness when applied to diverse datasets while effectively preventing overfitting to the specific characteristics of the proxy dataset used for calibration.

%% file: tables/in_distribution.tex
\renewcommand{\arraystretch}{1.0}
\begin{table}[t]
\caption{Comparison of approaches on in-distribution datasets.}
\label{tab:main_in}
\centering
\rowcolors{2}{lightgray}{}
\setlength{\tabcolsep}{0.6em}{
\begin{tabular}{lccccccccc}
\toprule[0pt]
& \multicolumn{3}{c}{CIFAR-100} & \multicolumn{3}{c}{ImageNet-R} \\
\cmidrule(l{10pt}r{10pt}){2-4}\cmidrule(l{10pt}r{10pt}){5-7}
Distribution $\beta$ & 1.0 & 0.5 & 0.2 & 1.0 & 0.5 & 0.2 \\
\midrule
FedAvg & 93.2 & 92.93 & 91.56 & 84.38 & 81.68 & 78.73\\
RegMean & 93.36 & 93.09 & 92.11 & 84.12 & 83.35 & 81.18\\
FisherAvg & \underline{93.4} & 93.34 & 92.22 & 84 & 83.78 & 82.67\\
FedProto & 92.99 & 92.9 & 89.88 & 84.27 & 82.2 & 76.97\\
CCVR & 93.27 & \underline{93.38} & \underline{92.35} & \underline{84.87} & \underline{84.17} & \underline{82.8}\\
\methname & \textbf{93.57} & \textbf{93.58} & \textbf{92.92} & \textbf{85.45} & \textbf{84.35} & \textbf{83.05}\\
\bottomrule
\end{tabular}}
\end{table}

%% file: tables/out_distribution.tex
\renewcommand{\arraystretch}{1.0}
\begin{table}[t]
\caption{Comparison of approaches on out-of-distribution datasets.}
\label{tab:main_out}
\centering
\rowcolors{2}{lightgray}{}
\setlength{\tabcolsep}{0.6em}{
\begin{tabular}{lccccccccc}
\toprule[0pt]
& \multicolumn{3}{c}{EuroSAT} & \multicolumn{3}{c}{CUB-200} \\
\cmidrule(l{10pt}r{10pt}){2-4}\cmidrule(l{10pt}r{10pt}){5-7}
Distribution $\beta$ & 1.0 & 0.5 & 0.2 & 1.0 & 0.5 & 0.2 \\
\midrule
FedAvg & 98.35 & \textbf{98.36} & 97.43 & 73.51 & 78.01 & 61.98\\
RegMean & 98.44 & \underline{98.28} & 97.47 & 74.85 & 70.94 & 59.8\\
FisherAvg & \textbf{98.49} & 97.98 & 96.86 & 73.11 & 75.16 & 75.92\\
FedProto & 98.31 & 98.15 & 97.11 & 79.05 & 74.51 & 62.79\\
CCVR & 98.44 & \textbf{98.36} & \textbf{97.83} & \underline{85.42} & \underline{84.79} & \textbf{84.66}\\
\methname & \underline{98.47} & 98.17 & \underline{97.53} & \textbf{86.24} & \textbf{85.73} & \underline{82.86}\\
\bottomrule
\end{tabular}}
\end{table}

%% file: tables/ablations.tex
\renewcommand{\arraystretch}{1.0}
\begin{table}[t]
\begin{minipage}[t]{0.536\textwidth}
    \caption{Evaluation of the combined performance of \methname and CCVR~\cite{luo2021no} on CUB-200 across varying $\beta$-distribution parameter.}
    \label{tab:composability}
    \centering
    \rowcolors{2}{lightgray}{}
    \setlength{\tabcolsep}{0.5em}{
    \begin{tabular}{lccc}
    \toprule[0pt]
    Distribution $\beta$ & 1.0 & 0.5 & 0.2 \\
    \midrule
    CCVR & 85.42 & 84.79 & 84.66\\
    \methname & 86.24 & 85.73 & 82.86\\
    \methname \plus  CCVR & \textbf{86.52} & \textbf{86.42} & \textbf{86.37}\\
    \bottomrule
    \end{tabular}}
\end{minipage}
\hfill
\begin{minipage}[t]{0.36\textwidth}
    \caption{Ablation study on \methname components.}
    \label{tab:ablation_components}
    \centering
    \rowcolors{2}{lightgray}{}
    \setlength{\tabcolsep}{0.66em}{
    \begin{tabular}{cccc}
    $\alpha^l$ & $\sigma^{(c)}$ & Avg (std) \\
    \midrule
    \xmark & \xmark & 79.16 (0.16) \\
    \cmark & \xmark & 79.81 (0.12) \\
    \xmark & \cmark & 90.63 (0.07) \\
    \cmark & \cmark & 90.90 (0.07) \\
    \bottomrule
    \end{tabular}}
    \end{minipage}
\end{table}

%% file: figures/shap_imagenetr.tex
\begin{figure}[t]
    \centering
    \includegraphics[width=\columnwidth]{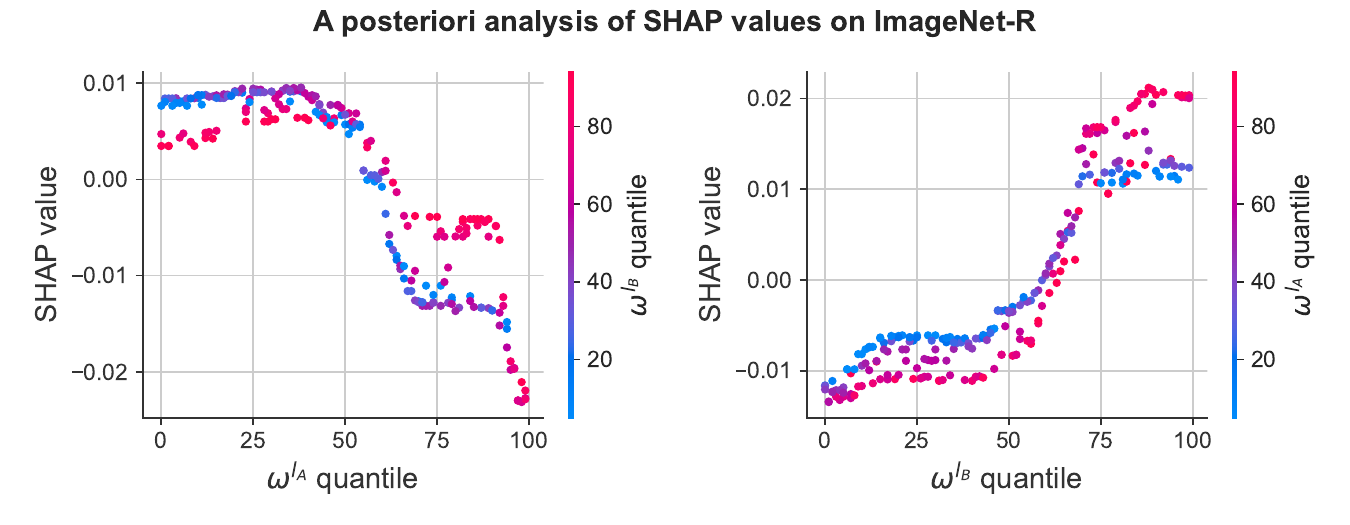}
    \caption{A posteriori analysis on $\omega_m^{l_A}$ and $\omega_m^{l_B}$ impact in predicting $\lambda_m^l$ coefficients on ImageNet-R. The trend of both features' impact is consistent with the one observed in the proxy dataset.}
    \label{fig:shapimgr}
\end{figure}

%% file: sections/6_conclusions.tex
\section{Conclusions}
We introduce \methname, a Federated Learning framework that advances model aggregation through two novel strategies. For backbone adaptation, we develop an explainability-driven merging procedure for LoRA modules based on training-emerging signals, while our classifier aggregation employs feature variance to weight client contributions. The method demonstrates state-of-the-art performance across multiple benchmarks while maintaining seamless compatibility with existing approaches.

Looking ahead, we identify promising opportunities to extend this work through local explainability techniques for coefficient estimation, potentially removing the need for proxy datasets and enabling more dataset-specific merging. We believe these directions will not only benefit Federated Learning but could also influence broader model fusion research, particularly for explainable model adaptation.

%% file: supplementary/0_supplementary.tex
\appendix
\renewcommand{\thetable}{\Alph{table}}
\renewcommand{\thefigure}{\Alph{figure}}

\setcounter{figure}{0}
\setcounter{table}{0}

%% file: supplementary/1_shapvalues.tex
\section{Details on $G_{SHAP}(\omega^{l_A},\omega^{l_B})$}
\label{sec:calbrationdetails}
\input{tables/shapcalibration}
The merging coefficients $\alpha_m^l$, detailed in the main paper, are derived through a discretization procedure applied to the importance signals $\omega_m^{l_A}$ and $\omega_m^{l_B}$. To mitigate overfitting on the proxy dataset, a discretization of these values is performed instead of directly utilizing the SHAP-explained regression model's predictions.

Specifically, these values are obtained by summing a SHAP base value of $0.063$ with the SHAP values of individual features, the latter being approximated through analysis of the SHAP proxy dataset figure presented in the main paper.

At the conclusion of each round, the server calculates the quantile values $Q_A$ and $Q_B$ for the complete distribution of $\omega_m^{l_A}$ and $\omega_m^{l_B}$, respectively, across all clients $m$ and layers $l$. Subsequently, for each layer, the corresponding value is assigned based on the intervals specified in \Cref{tab:shap_cal}, and a normalization is applied across each layer to ensure that the sum of the resulting SHAP values after the discretization of $\omega_m^{l_A}$ and $\omega_m^{l_B}$ equals $1$. This procedure defines the function $G_{SHAP}(\cdot)$, which is executed server-side at the end of each communication round, returning $\alpha^l$ coefficients.

Conceptually, this approach rewards clients exhibiting high $\omega^{l_B}$ and low $\omega^{l_A}$ values, assigns moderate weights to clients with either high levels of both $\omega^{l_B}$ and $\omega^{l_A}$ or intermediate $\omega^{l_B}$ levels, and assigns minimal weights to clients with low $\omega^{l_B}$ values.

%% file: tables/shapcalibration.tex
\renewcommand{\arraystretch}{2.0}
\begin{table}[th]
\caption{Calibrated values for $G_{SHAP}(\omega^{l_A},\omega^{l_B})$ based on $\omega^{l_A}$ and $\omega^{l_B}$ quantiles.}
\label{tab:shap_cal}
\centering
\setlength{\tabcolsep}{0.8em}{
\begin{tabular}{lccc}
& \textbf{$Q_B < 60$} & \textbf{$60 \le Q_B \le 80$} & \textbf{$Q_B > 80$} \\
\midrule
\textbf{$Q_A < 25$} & 0.038 & 0.100 & 0.193 \\
\rowcolor{lightgray}
\textbf{$25 \le Q_A \le 50$} & 0.038 & 0.089 & 0.118 \\
\textbf{$Q_A > 50$} & 0.038 & 0.078 & 0.093 \\
\bottomrule
\end{tabular}}
\end{table}

%% file: supplementary/2_varimportance.tex
\section{Robustness in the Choice of  $\omega_m^{l_A}$ and $\omega_m^{l_B}$}
\label{sec:omega}
\input{figures/var_importance}
The selection of $\omega_m^{l_A}$ and $\omega_m^{l_B}$ as predictors of the optimal weighting coefficients $\lambda_{m}^l$ stems from the intuition that a layer's contribution to loss variation can effectively capture its importance during server-side aggregation. We further investigate whether the number of examples seen during training could provide additional useful information for predicting the optimal coefficients.

Specifically, we initially trained the Random Forest using three features: $\omega_m^{l_A}$, $\omega_m^{l_B}$, and the number of training samples for each client. When evaluating the SHAP values of these three features, it became evident that the impact of the number of samples is negligible, as shown in \Cref{fig:varimp}. This result is directly attributable to the fact that $\lambda_{m}^l$ coefficients are estimated layer-wise. More importantly, this finding supports our hypothesis that $\omega_m^{l_A}$ and $\omega_m^{l_B}$ are sufficient and appropriate choices for predicting optimal merging coefficients.

%% file: figures/var_importance.tex
\begin{figure}[t]
    \centering
    \includegraphics[width=0.6\columnwidth]{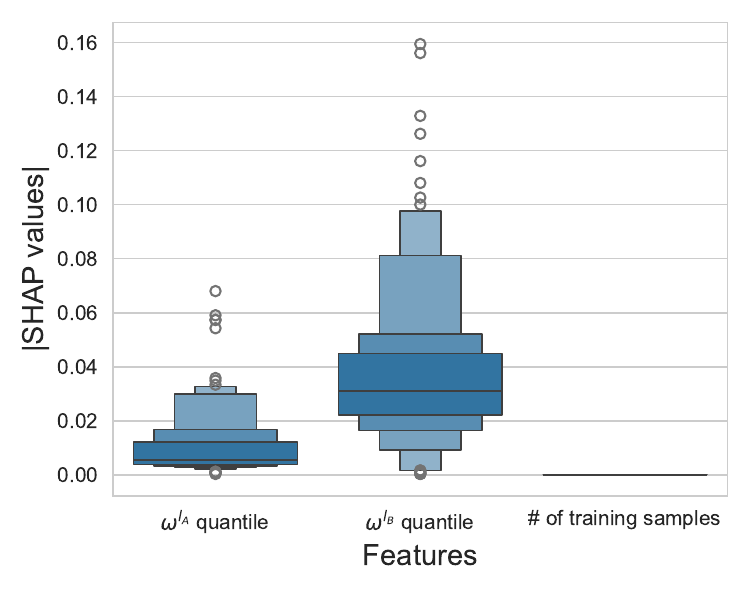}
    \caption{Evaluation of efficacy of the number of training samples in $\lambda_m^l$ prediction.}
    \label{fig:varimp}
\end{figure}

%% file: supplementary/3_a_posteriori.tex
\section{Additional a posteriori Analysis of SHAP Results}
\label{sec:additional_a_posteriori}
As demonstrated in the main paper, the SHAP trend observed for the $\omega^{l_A}$ and $\omega^{l_B}$ quantiles is consistent between the Tiny-ImageNet proxy dataset and other distributions. Figures \cref{fig:shapcifar}, \cref{fig:shapeur}, and \cref{fig:shapcub} further illustrate that this trend is consistently maintained across the CIFAR-100, EuroSAT, and CUB-200 datasets. Specifically, for CIFAR-100, the negative impact of $\omega^{l_A}$ at high quantiles is more pronounced at the upper extreme of the distribution. The impact of $\omega^{l_B}$ is generally similar across the datasets examined, with the notable exception of CUB-200, where its relative impact is lower in magnitude. Consequently, the non-decreasing trend for $\omega^{l_B}$ is not fully maintained across the entire distribution for this dataset. This observation is likely attributable to the dissimilarity between the CUB-200 dataset and the pretraining data, suggesting that the change in impact layer direction, represented by the $A$ matrices, exerts a more significant influence on the optimal merging strategy in this case.
\input{figures/shap_cifar}
\input{figures/shap_eurosat}
\input{figures/shap_cub200}

%% file: figures/shap_cifar.tex
\begin{figure}[t]
    \centering
    \includegraphics[width=\columnwidth]{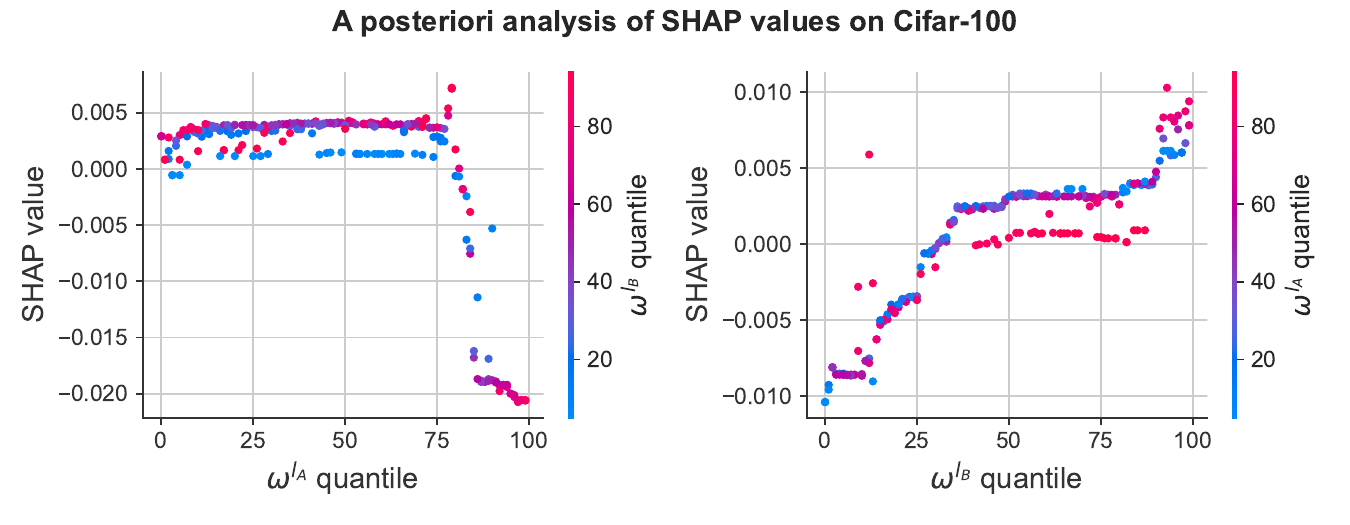}
    \caption{A posteriori analysis on $\omega_m^{l_A}$ and $\omega_m^{l_B}$ impact in predicting $\lambda_m^l$ coefficients on CIFAR-100.}
    \label{fig:shapcifar}
\end{figure}

%% file: figures/shap_eurosat.tex
\begin{figure}[t]
    \centering
    \includegraphics[width=\columnwidth]{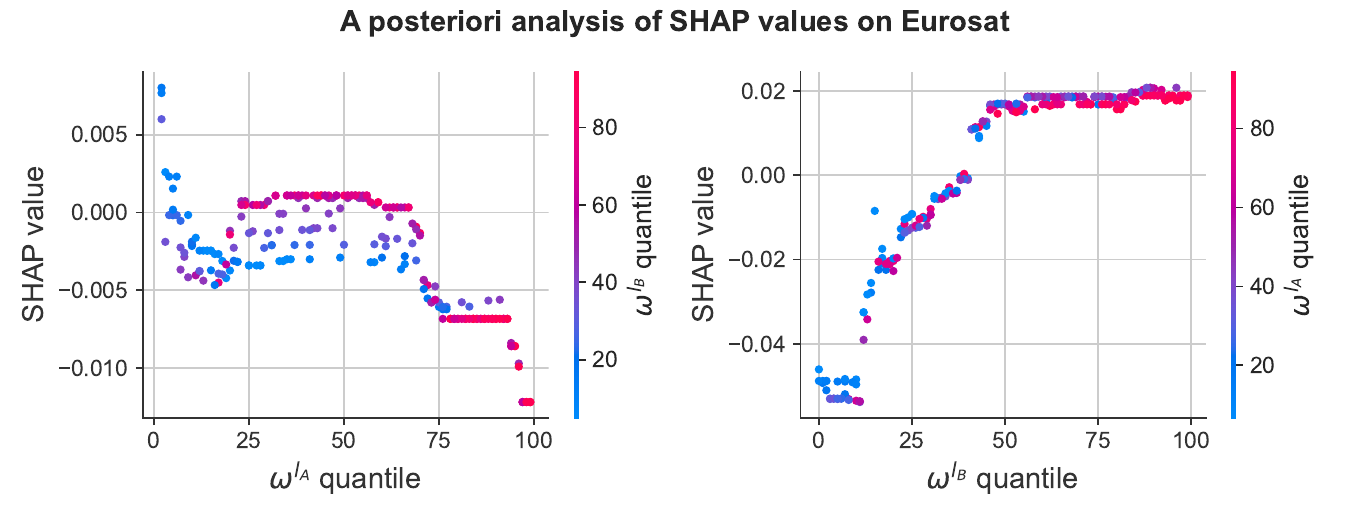}
    \caption{A posteriori analysis on $\omega_m^{l_A}$ and $\omega_m^{l_B}$ impact in predicting $\lambda_m^l$ coefficients on EuroSAT.}
    \label{fig:shapeur}
\end{figure}

%% file: figures/shap_cub200.tex
\begin{figure}[t]
    \centering
    \includegraphics[width=\columnwidth]{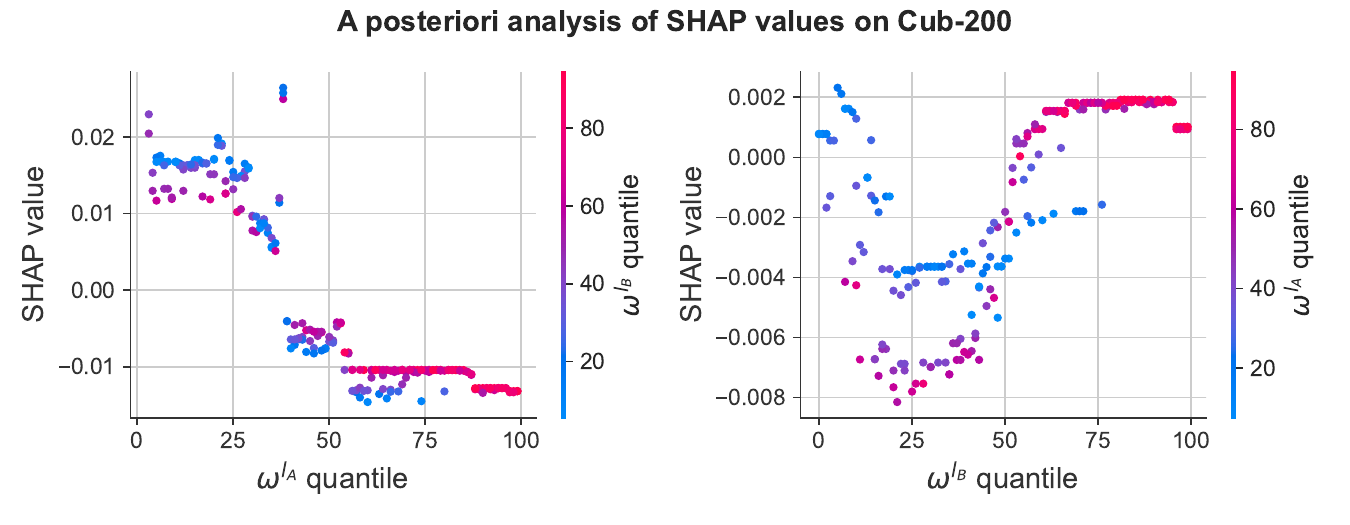}
    \caption{A posteriori analysis on $\omega_m^{l_A}$ and $\omega_m^{l_B}$ impact in predicting $\lambda_m^l$ coefficients on CUB-200.}
    \label{fig:shapcub}
\end{figure}